\documentclass{article}

\PassOptionsToPackage{numbers, compress}{natbib}
\usepackage[final]{nips_2017}

\usepackage[utf8]{inputenc} 
\usepackage[T1]{fontenc}    
\usepackage{hyperref}       
\usepackage{url}            
\usepackage{booktabs}       
\usepackage{amsfonts}       
\usepackage{microtype}      
\usepackage{graphicx}
\usepackage{natbib}
\usepackage{amsmath}
\usepackage{subcaption}

\title{Generating and designing DNA with deep generative models}

\author{
  Nathan Killoran \\
  University of Toronto\\
  \texttt{nkilloran@psi.toronto.edu} \\
  \And
  Leo J. Lee \\
  University of Toronto\\
  \texttt{ljlee@psi.toronto.edu} \\
  \And
  Andrew Delong \\
  University of Toronto\\
  \texttt{andrew.delong@gmail.com} \\
  \And
  David Duvenaud \\
  University of Toronto \\
  \texttt{duvenaud@cs.toronto.edu} \\
  \And
  Brendan J. Frey \\
  University of Toronto \\
  \texttt{frey@psi.toronto.edu} \\
}

\begin{document}

\maketitle

\begin{abstract}
  We propose generative neural network methods to generate DNA sequences and tune them to have desired properties. We present three approaches: creating synthetic DNA sequences using a generative adversarial network (GAN); a DNA-based variant of the activation maximization (``deep dream'') design method; and a joint procedure which combines these two approaches together. We show that these tools capture important structures of the data and, when applied to designing probes for protein binding microarrays (PBMs), allow us to generate new sequences whose properties are estimated to be superior to those found in the training data. We believe that these results open the door for applying deep generative models to advance genomics research.
\end{abstract}

\section{Introduction}

A major trend in deep learning
is the development of new generative methods, with the goal of creating synthetic data with desired structures and properties. 
This trend includes generative models such as generative adversarial networks (GANs) \cite{goodfellow2014generative}, variational autoencoders (VAEs) \cite{kingma2013auto}, 
and deep autoregressive models \cite{oord2016pixel, van2016conditional}, as well as generative design procedures like activation maximization (popularly known as ``deep dream'') \cite{simonyan2013deep, yosinski2015understanding, mordvintsev2015inceptionism} and style transfer \cite{gatys2015neural}.
These powerful generative tools bring many new opportunities. 
When data is costly, we can use a generative method to inexpensively 
simulate data. 
We can also use generative tools to explore the space of possible data configurations, tuning the generated data to have specific target properties, or to invent novel, unseen configurations which push beyond our existing understanding.
As well, these new methods can be applied to a variety of domains. 
As an example, Refs. \cite{gomez2016automatic, kadurin2017cornucopia, guimaraes2017objective, olivecrona2017molecular} have recently leveraged generative models to discover and design molecules which optimize certain chemically desired properties. 

In this work, we present deep generative models for DNA sequences. DNA sequence data could be considered as a hybrid between natural language data and computer vision data. Like in language, the data contains discrete sequences of characters (the nucleotides A, C, G, T) with some hierarchical structure (coding exon regions interspersed with non-coding intron regions). On the other hand, like in computer vision, there are regularly appearing patterns (motifs) placed on backgrounds that might be irrelevant or noisy. Thus, we hope to adapt design patterns from both of these established fields. 

There are many use-cases for generative tools in genomics. One candidate is automated design of the experimental probe sequences used to measure the binding of proteins to DNA or RNA. Another is the optimization of genomic sequences to satisfy multiple, possibly competing, properties (for instance, that the sequence must be translated by the cell into a specific protein product, while some other property, such as GC content, remains fixed). Or we could optimize genomic sequences so that a cell produces as much valuable chemical product (e.g., pharmaceuticals, biofuels) as possible. In the future, generative models could potentially serve as vital tools within the larger field of synthetic biology \cite{benner2005synthetic}.

Our overall goal will be to create synthetic DNA sequences and to tune these sequences to have certain desired properties. 
We present three complementary methods for achieving this goal: a GAN-based deep generative network for the creation of new DNA sequences; a variant of the activation maximization method for designing sequences with desired properties; and a joint method which combines these two components into one architecture. 
Through a number of computational experiments, we show that our generative model can capture underlying structure in a dataset and manifest this structure in its generated sequences. We also demonstrate that these methods are flexible and powerful tools for tuning DNA sequences to have specific properties. 
We have written the paper to be accessible to both machine learning researchers and computational biologists with interests in machine learning. 
We hope to inspire practitioners of both fields with these exciting new approaches.

\section{Generative Design of DNA}

We can picture generative methods falling in two main categories: generative modelling and generative optimization.
In this section, we will briefly review these two key methodologies. By convention, we convert discrete sequence data into a continuous representation using one-hot encodings, where each character in a sequence is encoded as a one-hot real vector, e.g., $A=[1,0,0,0]^T$, $C=[0,1,0,0]^T$, etc. A sequence $\mathbf{x}$ of length $L$ will thus be represented as a real-valued matrix of shape $(L, 4)$, i.e., $\mathbf{x}\in\mathbb{R}^{L}\times\mathbb{R}^4$.

\subsection{Generative Models}

In many domains of interest (e.g., images, natural language, genomics, etc.), the set of all possible data configurations (pixel values, character/word sequences, nucleotide sequences) can be prohibitively large. 
Yet we can recognize that many of these configurations, such as random noisy images or nonsensical character combinations, are irrelevant or unlikely in real use-cases. 
Plausible real-world examplars occupy only a small structured subset of configurations. 
The goal of a generative model is to capture this underlying realistic structure.


A generative network, parameterized by some variables $\theta_G$, will encode -- either explicitly or implicitly -- its own generative distribution $\mathcal{P}_{G}$ through which it produces samples of synthetic data. 
During training, we tune the parameters $\theta_G$ until the model's samples are similar to real-world examples, i.e., $\mathcal{P}_G\approx\mathcal{P}_\mathrm{real}$. This is accomplished using a dataset containing typical examples $\mathbf{x}\sim\mathcal{P}_\mathrm{real}$. The network is trained to generate samples similar to this dataset and hence also to true data.

In the past, we might have hand-crafted the generative process $\mathcal{P}_G$ using subject manner expertise. For example, in Ref. \cite{shai2010machine}, a probabilistic model was constructed to generate DNA sequences in which certain motifs (i.e., subsequences which follow specific statistical patterns) were present. In contrast, modern deep learning methods allow us to train large flexible models end-to-end from data without costly expert construction. 

There are a number of architecture choices available for generating genomic sequences with neural networks. We have experimented with recurrent neural networks, the PixelCNN model, VAEs, and GANs; more discussion and observations on these can be found in Appendix \ref{app:gen_archs}. In the remainder of the main text, we make use of a GAN-based generative model, which we will outline next.

\subsubsection{Generative Adversarial Networks}

\begin{figure}[t]
 \centering\includegraphics[width=.9\textwidth]{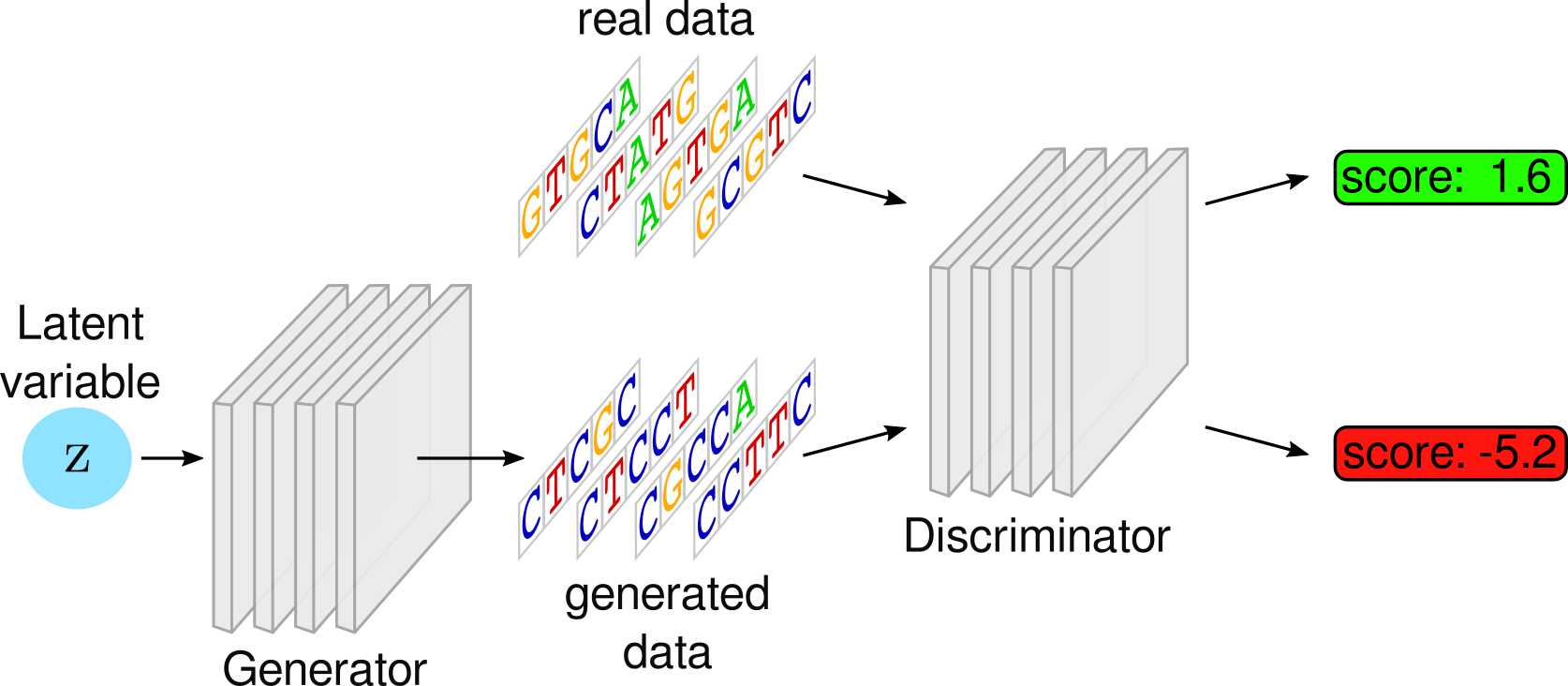}
 \caption{Basic schematic of a Wasserstein Generative Adversarial Network (WGAN) architecture. The generator and discriminator are both parameterized as deep neural networks. For clarity, short exact data sequences are shown. In practice, the real data is encoded using one-hot vectors, and the generator learns a continuous approximation to this encoding.}\label{fig:gan_outline}
\end{figure}

GANs contain two components (Fig. \ref{fig:gan_outline}). First, a generator $G$ transforms a continuous variable $\mathbf{z}$ into synthetic data, $G(\mathbf{z})$. The variable $\mathbf{z}$ serves as a high-level latent encoding for the data. The second component is a discriminator $D$, whose role is to distinguish generated data from real data. $D$ receives batches of data $\mathbf{x}$ (either real or generated) and produces a single real number output $D(\mathbf{x})$. In the original GAN formulation \cite{goodfellow2014generative}, $D(\mathbf{x})$ was bounded in $[0,1]$, representing the probability that the data is real. In the Wasserstein GAN (WGAN) reformulation \cite{arjovsky2017wasserstein, gulrajani2017improved}, the discriminator's output is adapted to an arbitrary score $D(\mathbf{x})\in\mathbb{R}$, and an optimization penalty is introduced to bound the discriminator's gradients, making the model more stable and easier to train. In this work, we use a WGAN.

Training consists of alternating phases of discriminator and generator training. The discriminator is trained using both real data and synthetic data. The real data is taken from some real-world distribution or dataset $\mathcal{P}_\mathrm{real}$. The synthetic data is obtained by sampling the latent variable $\mathbf{z}$ according to some simple prior distribution $\mathcal{P}_\mathbf{z}$ (here we use a standard normal), and passing these through the generator network. The discriminator's training objective is, without knowing where the data came from, to maximize the scores of real data while minimizing the scores of fake data, i.e.,
\begin{equation}
 \max_{\theta_D} \mathcal{L}_\mathrm{disc} = \max_{\theta_D}~[\mathbb{E}_{\mathbf{x}\sim \mathcal{P}_\mathrm{real}}D(\mathbf{x}) - \mathbb{E}_{\mathbf{z}\sim \mathcal{P}_\mathbf{z}} D(G(\mathbf{z}))]
\end{equation}
where $\theta_D$ are the discriminator's model parameters. 
When training the generator, fake data samples are generated and passed through to the discriminator to be scored. The generator training objective is to maximize those scores:
\begin{equation}
 \max_{\theta_G} \mathcal{L}_\mathrm{gen} = \max_{\theta_G}~\mathbb{E}_{\mathbf{z}\sim \mathcal{P}_\mathbf{z}} D(G(\mathbf{z})).
\end{equation}
Both the generator and discriminator can be trained via standard gradient descent algorithms. When the discriminator learns to distinguish fake data from real data, it provides a meaningful learning signal for the generator. 
In order to ensure this, we can perform $N$ discriminator updates for each generator update. After training, we use just the generator network to create synthetic data.

GANs have shown rapid and exciting progress in continuous domains such as images, but reliably generating discrete sequences has posed major challenges. Only recently have GAN models been proposed which can successfully generate nontrivial natural language sequences \cite{yu2017seqgan, che2017maximum, hjelm2017boundary, gulrajani2017improved}, though these models can still be difficult to train reliably and are not yet state-of-the-art on natural language. A promising candidate was presented in Ref. \cite{gulrajani2017improved}, where a WGAN was trained to generate character-level English sentences with performance approaching that of standard recurrent neural network approaches. Our generative model for DNA will be based on this particular WGAN architecture. Further specific modelling details can be found in Appendix \ref{app:models}.

\subsection{Generative Optimization}

Instead of generating realistic-looking data, the focus in this alternative approach is to synthesize data which strongly manifests certain desired properties. We phrase this task as an optimization problem, rather than a modelling problem. Two well-known examples are the activation maximization \cite{simonyan2013deep, yosinski2015understanding, mordvintsev2015inceptionism} approach, where we use an image classifer network `in reverse,' to create images belonging to a particular class, and style transfer \cite{gatys2015neural}, where the content of one image is transformed to fit the style of another. We focus in this paper on two techniques: a version of activation maximization which works on discrete sequences like DNA; and a joint method which extends the activation maximization methodology by incorporating a trained generator model. 


\subsubsection{Activation Maximization for DNA}
\label{sec:deep_dream_for_dna}

Suppose we have a function $P$ which takes data as input and calculates or predicts some target property $\mathbf{t}=P(\mathbf{x})$. In computer vision, this property is usually the probability of an image belonging to a class of interest (e.g, \emph{dog}). However, we can consider $P$ in a more general manner. It can be a known, explicitly coded function $P(\mathbf{x})=f(\mathbf{x})$, a parameterized neural network model which is learned from data, $P(\mathbf{x})=f_\theta(\mathbf{x})$, or, most generally, some combination of explicit functions $\{f_i\}$ and learned functions $\{f_{\theta_j}\}$ which trades off multiple properties at once,
\begin{equation}\label{eq:multi_objective}
 P(\mathbf{x})=\sum_i \alpha_i f_i(\mathbf{x}) + \sum_j \beta_j f_{\theta_j}(\mathbf{x}),
\end{equation}
where $\{\alpha_i\}$ and $\{\beta_i\}$ are fixed weights indicating the relative influence of each property. For convenience, we will use the name `predictor' in all three situations. 

The basic activation maximization method works as follows. Starting with an arbitrary input, we calculate the gradient $\nabla_\mathbf{x}\mathbf{t}$, which tells us how the property $\mathbf{t}$ changes as we change $\mathbf{x}$. We take steps 
\begin{equation}
 \label{eq:deepdream_step_x}
 \mathbf{x} \rightarrow \mathbf{x} + \epsilon\nabla_\mathbf{x}\mathbf{t}
\end{equation}
along this gradient (where $\epsilon$ is a small step size), progressively modifying the original input $\mathbf{x}$ so that it increases or decreases $\mathbf{t}$ towards a desired value, or finds its optimal values.

While originally for images, this approach can also work on DNA sequences, provided we are careful about the discrete nature of $\mathbf{x}$. To this end, we obtain a continuous relaxation of one-hot sequence vectors by adding a simple unstructured latent variable $\mathbf{z}$. This latent variable has shape $(L,4)$, the same as the data encoding. Latent vectors are transformed into sequences via a simple softmax pre-layer: 
\begin{equation}
 \mathbf{x}_{ij} = \frac{\exp(\mathbf{z}_{ij})}{\sum_{k=1}^4 \exp(\mathbf{z}_{ik})}.
\end{equation}
This pre-layer ensures that at each position in a DNA sequence, we have a normalized probability distribution over the nucleotides. These smoothed-out encodings are then used as input to the predictor function $P$, and gradients are calculated with respect to the variable $\mathbf{z}$ rather than directly on the data. Our optimization steps will likewise take place in the continuous latent space: 
\begin{equation}
 \label{eq:deepdream_step_z}
 \mathbf{z} \rightarrow \mathbf{z} + \epsilon\nabla_\mathbf{z}\mathbf{t}.
\end{equation}
Once the optimization has finished, we can interpret the derived distributions as discrete sequences by taking the argmax at each position. 

\subsubsection{Joint method}
\label{sec:plug_and_play}

One drawback with the activation maximization approach is that it ignores realism or typicality of data in its pursuit of optimal attributes. For instance, such images are often exaggerated or nightmarish, with the target property manifesting in unrealistic ways (e.g., a generated image contains many disconnected dog faces). 

A clever approach to overcome this is to combine activation maximization with a generative model. This idea, termed ``plug \& play generative networks'', was proposed in \cite{nguyen2016synthesizing, nguyen2016plug} for the image domain, demonstrating impressive results. 
We can let a generator capture the generic high-level structure of data, while using predictors to fine-tune specific properties. Generator and predictor models can even be trained on different datasets (e.g., a large unlabelled dataset for the generator and a smaller labelled dataset for the predictor). A single predictor can also be paired with different generators, each trained to capture different types of genomic sequences, such as from functionally distinct regions or different species. Similarly, a single generator can be combined with multiple predictors. 

\begin{figure}[t]
 \centering\includegraphics[width=\textwidth]{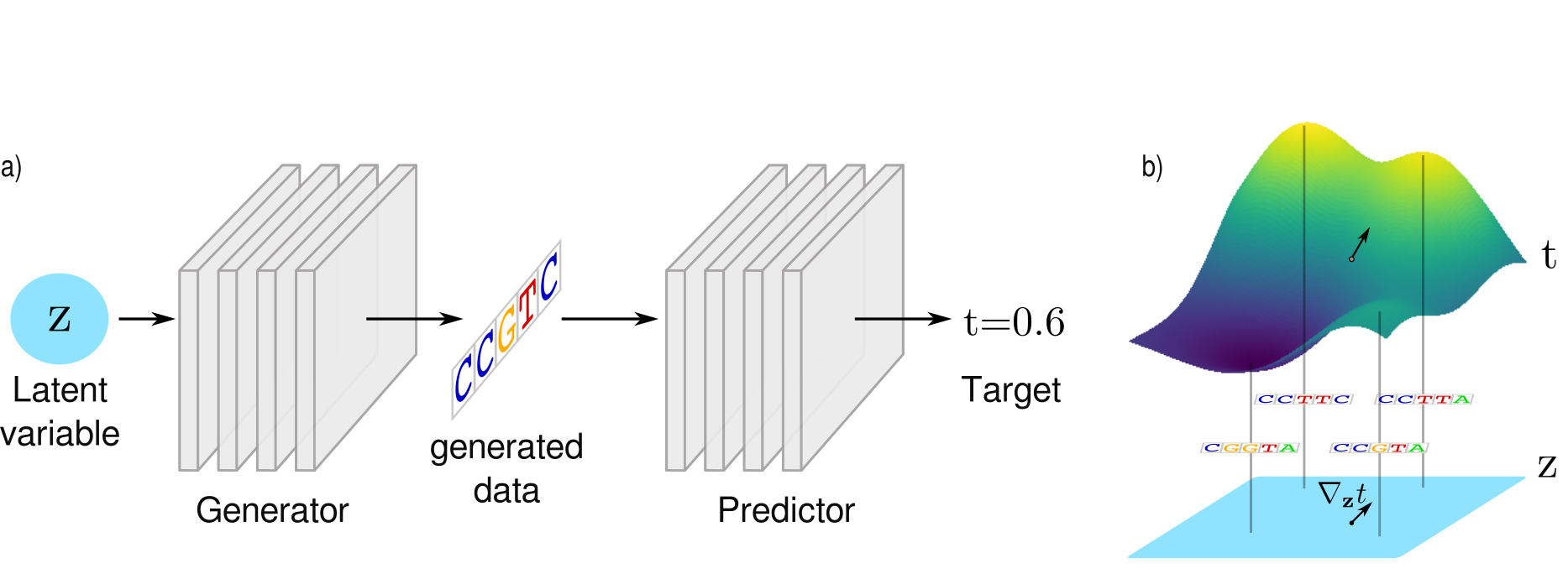}
 \caption{Joint architecture used for generating and tuning genomics sequences. 
 a) The generator transforms a latent variable $\mathbf{z}$ into synthetic data which is then passed through the predictor to compute the target $\mathbf{t}$. 
 The predictor can be a learned model (shown) or an explicit hard-coded function. Generator and predictors are built independently and different models can be mixed and matched.
 b) Schematic of DNA design process. Sequences can be tuned to have desired properties by calculating the gradient of $\mathbf{t}$ with respect to $\mathbf{z}$. We then take steps in the direction of the gradient in the latent space. As we move, the corresponding generated sequences (four examples shown) will change to ones which better manifest the desired properties.}\label{fig:plugandplay}
\end{figure}

This joint architecture requires two components: a generator $G$ which transforms latent codes $\mathbf{z}$ into synthetic data $\mathbf{x}$ (e.g., a trained GAN generator), and a predictor $P$, mapping data $\mathbf{x}$ to the corresponding target attributes $\mathbf{t}=P(\mathbf{x})$. The two modules are `plugged' back-to-back, so that they form a concatenated transformation $\mathbf{z}\rightarrow\mathbf{x}\rightarrow\mathbf{t}$. The combined architecture is shown in Fig. \ref{fig:plugandplay}.

In the joint method, the goal is still the same as activation maximization: tune data to have desired properties. To do this, we calculate the gradient of the prediction $\mathbf{t}$ with respect to the generator's latent codes $\mathbf{z}$, 
\begin{equation}\label{eq:grad_t_x}
 \nabla_\mathbf{z}\mathbf{t}
 =\sum_i\frac{\partial\mathbf{t}}{\partial\mathbf{x}_i}\frac{\partial\mathbf{x}_i}{\partial\mathbf{z}}
 =\sum_i\frac{\partial P(\mathbf{x})}{\partial\mathbf{x}_i}\frac{\partial G_i(\mathbf{z})}{\partial\mathbf{z}}, 
\end{equation}
where $i$ runs over all dimensions of the data. Once we have this, we can follow the same optimization recipe as Eq. (\ref{eq:deepdream_step_z}) above. 
Through this process, synthetic data is generated at the junction between the generator and the predictor which manifests the current attribute value. Note that we can consider activation maximization to be an architecture with a predictor but no generator (or, in our formulation for DNA above, a very simple generator). 


\section{Computational Experiments}

In the following, we will explore several computational experiments to test and validate our proposed DNA design methods. For each of these, further details about employed models and datasets can be found in Appendix \ref{app:models}.

\subsection{Generative DNA Model}

In this section, we perform several experiments intended to more fully understand the capabilities of our DNA generator architecture.  

\subsubsection{Exploring the Latent Encoding}
\label{sec:explore_latent}

A common method to explore and understand the structure of a generative architecture is to train a model on a large dataset and explore what it has learned via elementary manipulations of the latent space. To this end, we trained a WGAN model on a dataset of 4.6M 50-nucleotide-long sequences encompassing chromosome 1 of the human genome hg38 \cite{kent2002human}.


\begin{figure}[t]
 \centering\includegraphics[width=\textwidth]{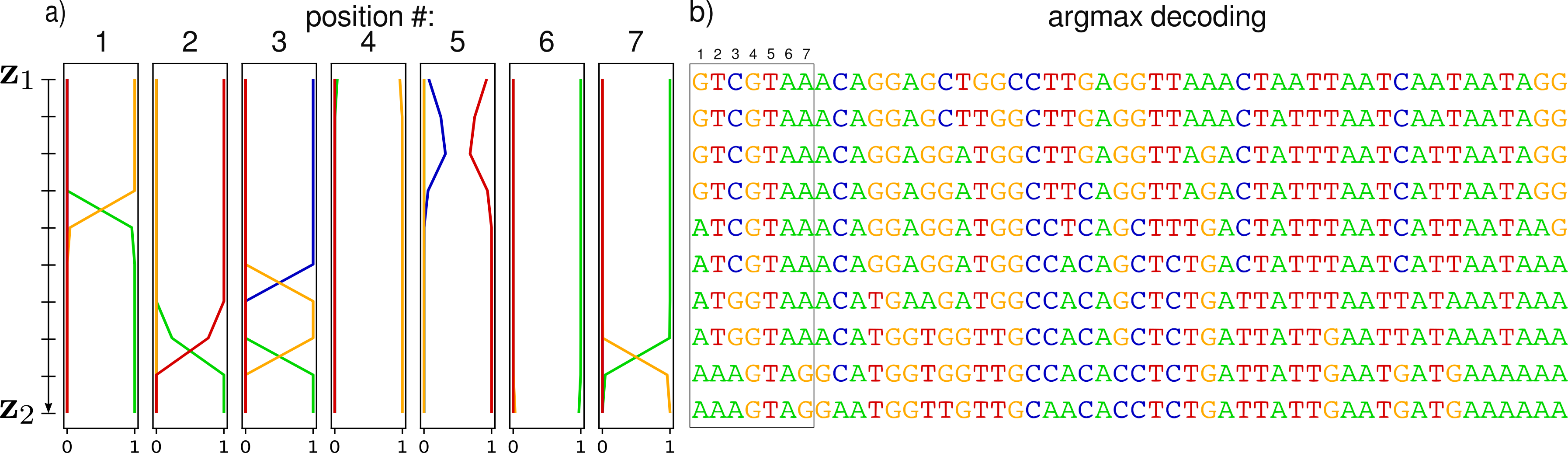}
 \caption{Linear interpolation between two randomly chosen points in latent space, $\mathbf{z}_1$ and $\mathbf{z}_2$. Latent values are interpolated top-to-bottom, and the corresponding generated sequences are shown left-to-right. 
 a) Generated data for the first seven positions in the output. Colours represent different output channels, one for each of the four characters A (green), C (blue), G (yellow), and T (red). 
 b) Corresponding full sequences decoded by taking the argmax over the four channels for each position in the sequence.}\label{fig:latent_interpolation}
\end{figure}

\paragraph{Latent interpolation} In our first exploration, we consider interpolation between points in the latent space.  In Fig. \ref{fig:latent_interpolation} we show how the generated data varies as we traverse a straight line between two arbitrary latent coordinates $\mathbf{z}_1$ and $\mathbf{z}_2.$ From Fig. \ref{fig:latent_interpolation}a, we can clearly see that the generator has learned to output a good approximation of one-hot data. For most of the latent points, one character channel has a value near one while the other three have values near zero. In regions where the decoded sequences change, the output switches smoothly between different one-hot vectors.

\begin{figure}[t]
 \centering\includegraphics[width=\textwidth]{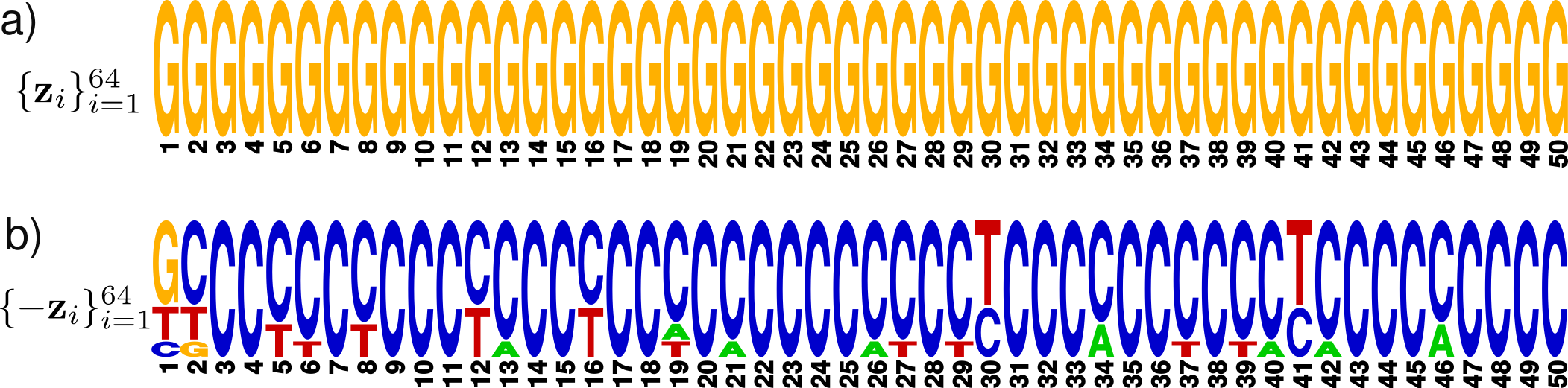}
 \caption{Latent space complementation with a batch of points $\{\mathbf{z}_i\}_{i=1}^{64}$. 
 a) Each latent code in the batch generates the same 50-nucleotide sequence (a sequence containing all G's). 
 b) The reflections of these latent codes, $\{-\mathbf{z}_i\}_{i=1}^{64}$, generate sequences strongly biased towards the complementary base C. Letter heights reflect their relative frequency in the batch at each position.}\label{fig:latent_complementation}
\end{figure}

\paragraph{Latent complementation} As a second exploration, we consider the effect of reflection in the latent space: $\mathbf{z}\rightarrow-\mathbf{z}$. To account for multiple latent vectors generating the same data, we fix a sequence $\mathbf{x}^*$ and find, via gradient-based search, 64 different latent points $\{\mathbf{z}_i\}_{i=1}^N$ which each generate $\mathbf{x}^*$, i.e., $G(\mathbf{z}_i)=\mathbf{x}^*$ for all $\mathbf{z}_i$. In this example, $\mathbf{x}^*$ will be the sequence with all G's. We reflect each of these latent points and decode the corresponding generated sequences, $\{\mathbf{x}_i^R\}_{i=1}^{64} = \{G(\mathbf{-z}_i)\}_{i=1}^{64}$. The results of this procedure are summarized in Fig. \ref{fig:latent_complementation}. 

Clearly, the reflected coordinates have a strong bias for generating C's. Readers may recognize that C is the complementary nucleotide to G, i.e., it is the only nucleotide which will form a base pair with G in stable double-helix DNA. Remarkably, the generative model seems to have learned the notion of DNA complementation from the training data and used negation in the latent space to approximately encode this structure. The model shows similar reflection structure for the other base-pair combinations (see Appendix \ref{app:complementation} for further discussion). 

\begin{figure}[t]
 \centering\includegraphics[width=0.7\textwidth]{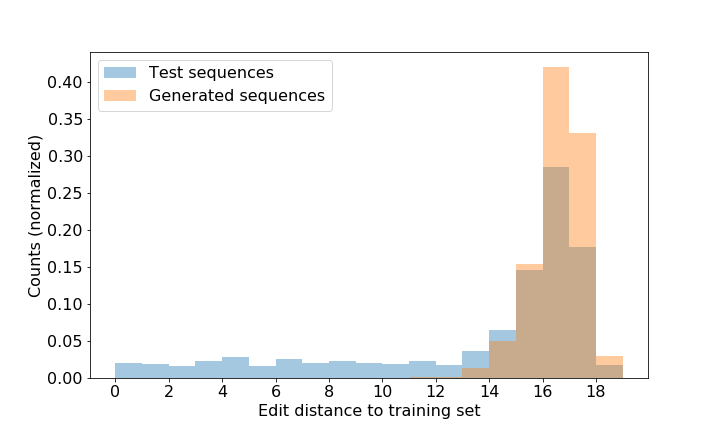}
 \caption{Edit distance of generated and test sequences to the training sequences.}\label{fig:edit_distance}
\end{figure}

\paragraph{Distance to training sequences} As a final check of the GAN model, we calculate the edit distance of generated sequences to the training set (Fig. \ref{fig:edit_distance}). For comparison, we also sample 1000 random sequences from a held-out test set and compute their edit distances to the training set. Both the generated and test sequences are concentrated around the same distance, but the generated sequences are, on average, further from the training sequences than the test sequences, indicating that the model has not overfit.

\subsubsection{Capturing Exon Splice Site Signals}
In another test of our GAN architecture, we trained a model on 116k 500-nt-long human genomic sequences, each containing exactly one exon (i.e., a structured coding subsequence which gets transcribed to mature RNA). The exons varied in length between 50-400 nucleotides. In order to recognize where the exons are located within the longer sequences, we included an additional position annotation track with the training data, such that nucleotide positions within an exon have a value of 1 and non-exon positions have a value of 0. Thus, the model must simultaneously learn to demarcate exons within the annotation track while also capturing the statistical information of nucleotides relative to these exon borders (so-called splice sites).

\begin{figure}[t]
 \centering\includegraphics[width=\textwidth]{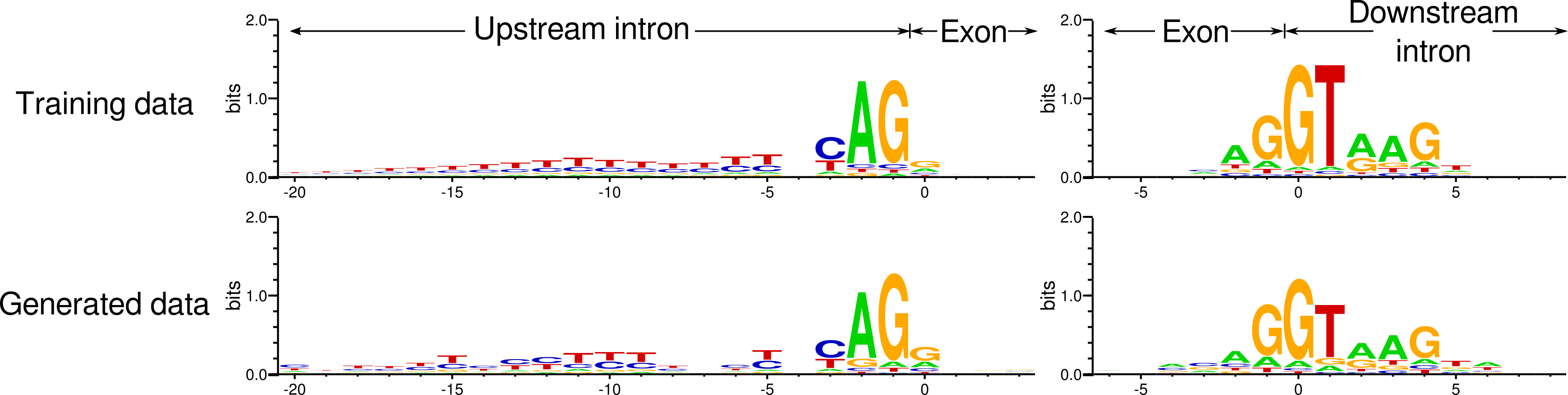}
 \caption{Exon splice site signal for the training data versus generated data. Exon starts/ends are located at position 0. Letter heights reflect the information content at each position.}\label{fig:exons}
\end{figure}

To inspect the results of our trained model, we used the generated annotation track to align the corresponding generated sequences (taking the first/last value above 0.5 as the start/end of the exon). Fig. \ref{fig:exons} shows the sequence logos for exon starts and exon ends, both for the original training data and for our model-generated data. We can see that the model has picked up on various splice site signals and its generated sequences approximate these signals fairly well. This experiment hints at an advantage of the GAN framework; based on the results here and extrapolating results from computer vision, GANs can potentially scale up to large gene-scale sequences (thousands of nucleotides or more), providing a new tool for the design of whole genes, and perhaps one day even small genomes.

\subsection{Designing DNA}

In this section, we undertake several experiments for designing DNA sequences. Our running theme will be DNA/protein binding. As we progress, the design task will become more and more challenging. We first optimize an explicit, hand-coded predictor which encodes a known motif. Then we introduce a learned binding predictor and try to automatically generate new high-binding-affinity sequences. Finally, we attempt to design sequences which must trade off two binding affinities at once. 

\subsubsection{Explicit Predictor: Motif Matching}
\label{sec:motif_matching}

\begin{figure}[t]
 \centering\includegraphics[width=\textwidth]{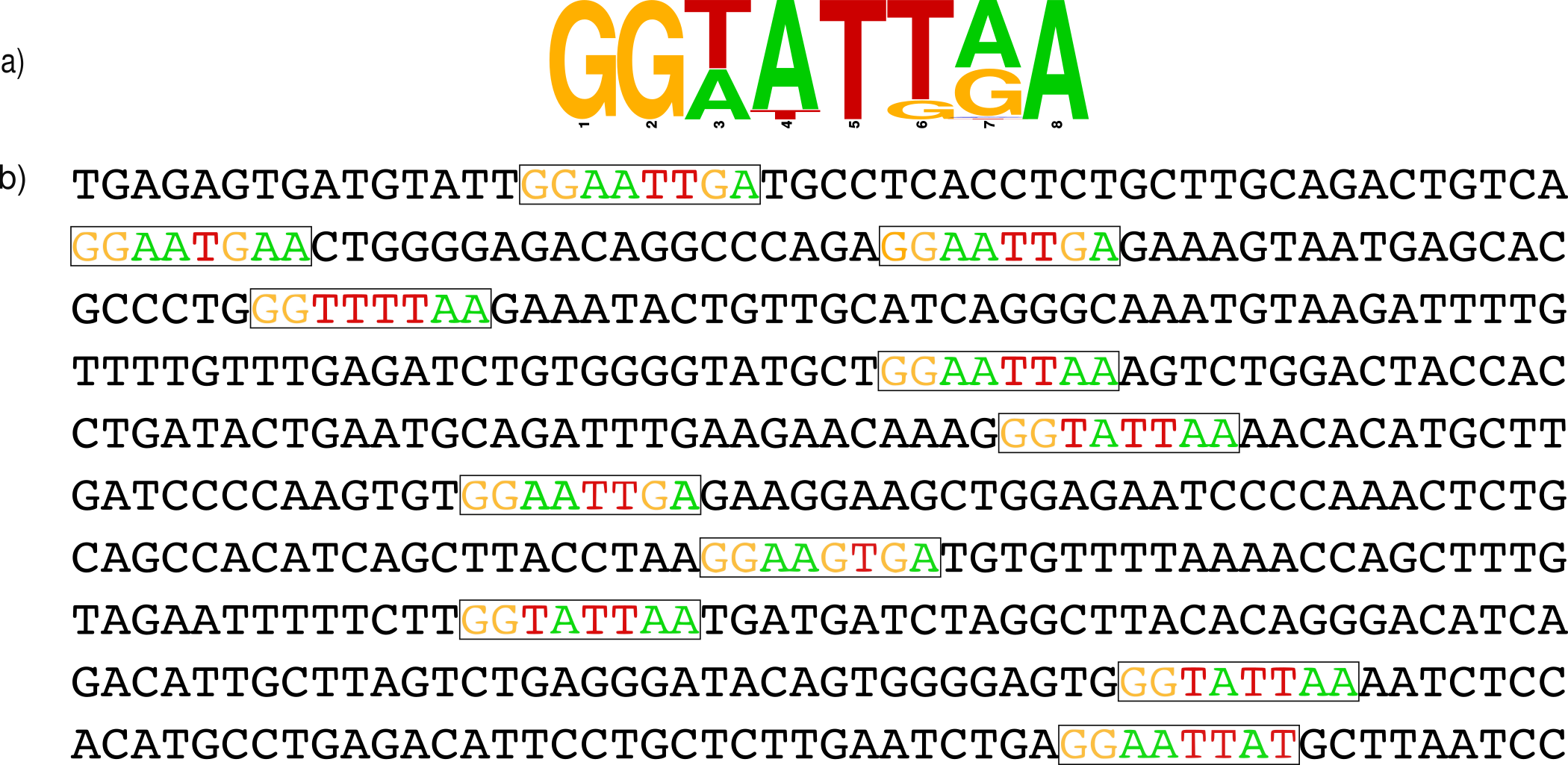}
 \caption{Motif-matching experiment: 
 a) Sequence logo for the PWM detected by the predictor. Letter heights reflect their relative frequency at each position. Sequences which have a strong match with this motif will score highly.
 b) Sample sequences tuned to have a high predictor score. The boxes indicate strong motif matches for each sequence.}\label{fig:motif_matching}
\end{figure}

Here our goal is to design DNA sequences using an explicit biologically motivated predictor function. 
In genomics data, there are short subsequences, or \emph{motifs}, which appear with much more regularity than expected by random chance. These motifs often serve functional roles, for instance as binding sites for cellular proteins. Motifs are often soft, in the sense that there is some flexibility about which nucleotide appears in a given position. Mathematically, a motif of length $K$ is commonly represented by a $K\times 4$ position weight matrix (PWM), where each column forms a probability distribution representing the relative frequency of each nucleotide at the corresponding position. Graphically, motifs can be represented with a sequence logo like the one shown in Fig. \ref{fig:motif_matching}a. 

\paragraph{Design process} Our predictor function consists of two stages. The first is a one-dimensional convolution which scans across the data, computing the inner product of a \emph{fixed} PWM (analogous to a convolutional filter from computer vision) with every length-$K$ subsequence of the data. We then select the single convolutional output with the highest value (`max-pooling') to get the final score for the sequence. A sequence will have high score as long as it has a single subsequence which has a large inner product with the chosen PWM. For this experiment, we used the joint method, employing a generator trained on sequences from human chromosome 1, as in Sec. \ref{sec:explore_latent}. Sample results of this optimization are shown in Fig. \ref{fig:motif_matching}b, demonstrating that we were successful in generating sequences that manifest the desired motif.

\subsubsection{Learned Predictor: Protein Binding}
\label{sec:learned_predictor}

In this experiment, we explore the use of a predictor model which has been learned from data. Specifically, we will focus on data from protein-binding experiments. 
Our dataset for this and the following experiment was provided by \cite{raluca2017}. It consists of DNA sequences and experimentally measured binding affinities of various transcription factor proteins, reflecting how strongly the given proteins bind to a probe containing that sequence. Our goal is to design new sequences which have high binding scores. 

\begin{figure}[t]
 \centering\includegraphics[width=\textwidth]{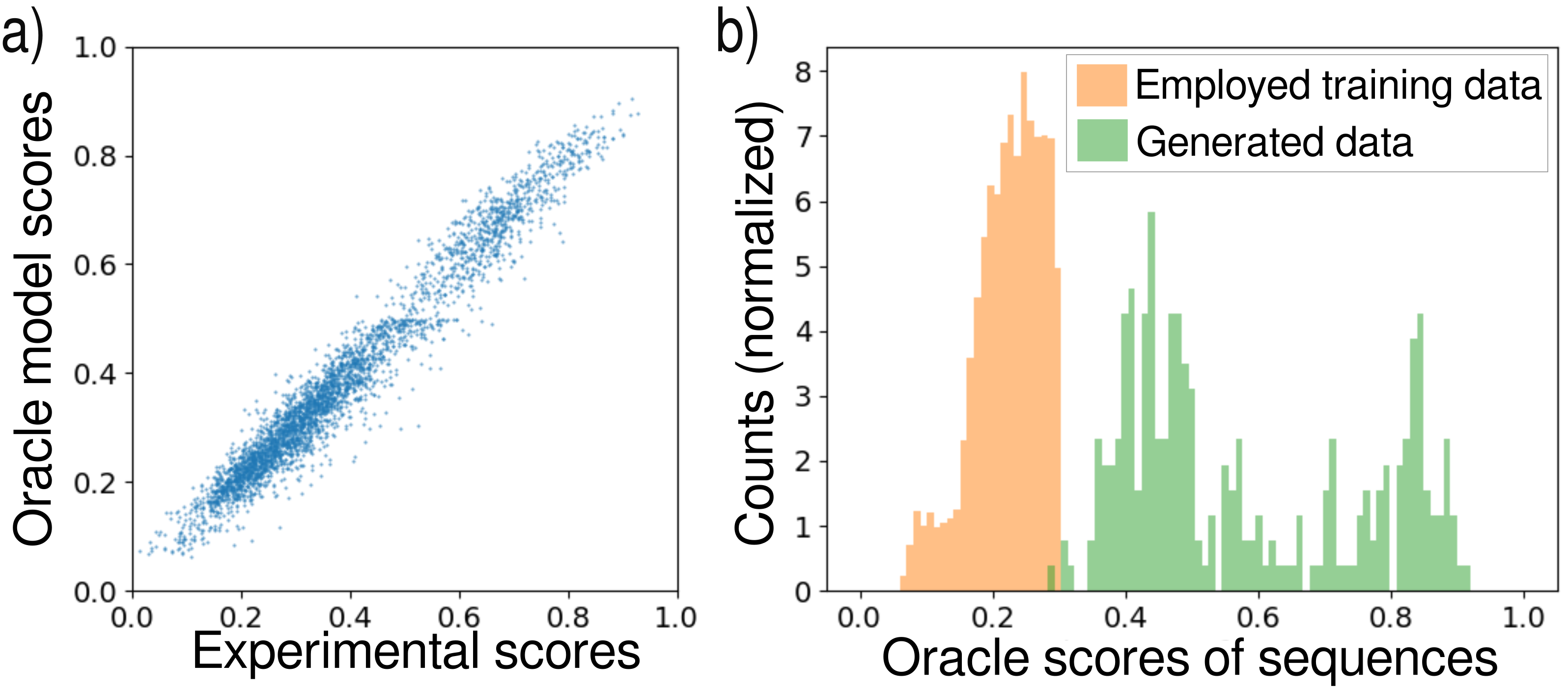}
 \caption{Protein binding optimization with a learned predictor model. 
 a) Original experimental data contains sequences and measured binding scores (horizontal axis); we fit a model to this data (vertical axis) to serve as an oracle for scoring generated sequences. Plot shows scores on held-out test data (Spearman correlation 0.97).
 b) Data is restricted to sequences with oracle scores in the 40th percentile (orange distribution), then used to train a generator and predictor model. Generated sequences are optimized to have as high binding score as possible. These genererated samples are then scored with the oracle (green distribution). The design process has clearly picked up enough structure that it can generalize well beyond the training data.}\label{fig:protein_binding}
\end{figure}

\paragraph{Oracle model} To simulate the process of evaluating candidate sequences, we will use a proxy model which is trained on the original experimental dataset (see Sec. \ref{app:models} for model details). This model will serve as an independent oracle which we can query with new designed sequences to gauge their expected binding score. For each sequence in the experimental data, we evaluate the binding score given by the oracle model. A comparison of experimentally measured versus oracle-predicted binding scores is found in Fig. \ref{fig:protein_binding}a, showing a strong correlation. 

\paragraph{Design process} For the DNA design task, our training data consisted of the original experimental probe sequences, but with the oracle-based scores.
To make the design task more challenging and to test generalization, we removed all data with scores above the 40th percentile and trained both a predictor and a generator on this restricted dataset. To emphasize, \emph{neither model saw any scores beyond the 40th percentile.}
Nevertheless, as can be seen in Fig. \ref{fig:protein_binding}, after optimization using our joint method, the designed sequences nearly all have scores higher than anything seen in the training set. Some designed sequences even have binding values three times higher than anything in the training data. This result indicates that a generative DNA design approach can be quite powerful for designing probe sequences even when only a weak binding signal is available.

\subsubsection{Optimizing Multiple Properties}
\label{sec:multi_objective}

As noted in Sec. \ref{sec:deep_dream_for_dna}, the activation maximization method can be used to simultaneously optimize multiple -- possibly competing -- properties. The joint method already does this to some extent. The predictor directs generated data to more desirable configurations; at the same time, the generator constrains generated data to be realistic. In this experiment, we performed a simultaneous activation maximization procedure on two predictors, each computing a different binding score. While we do not employ a generator, in principle one could also be included.

\begin{figure*}[t!]
    \centering\includegraphics[width=\textwidth]{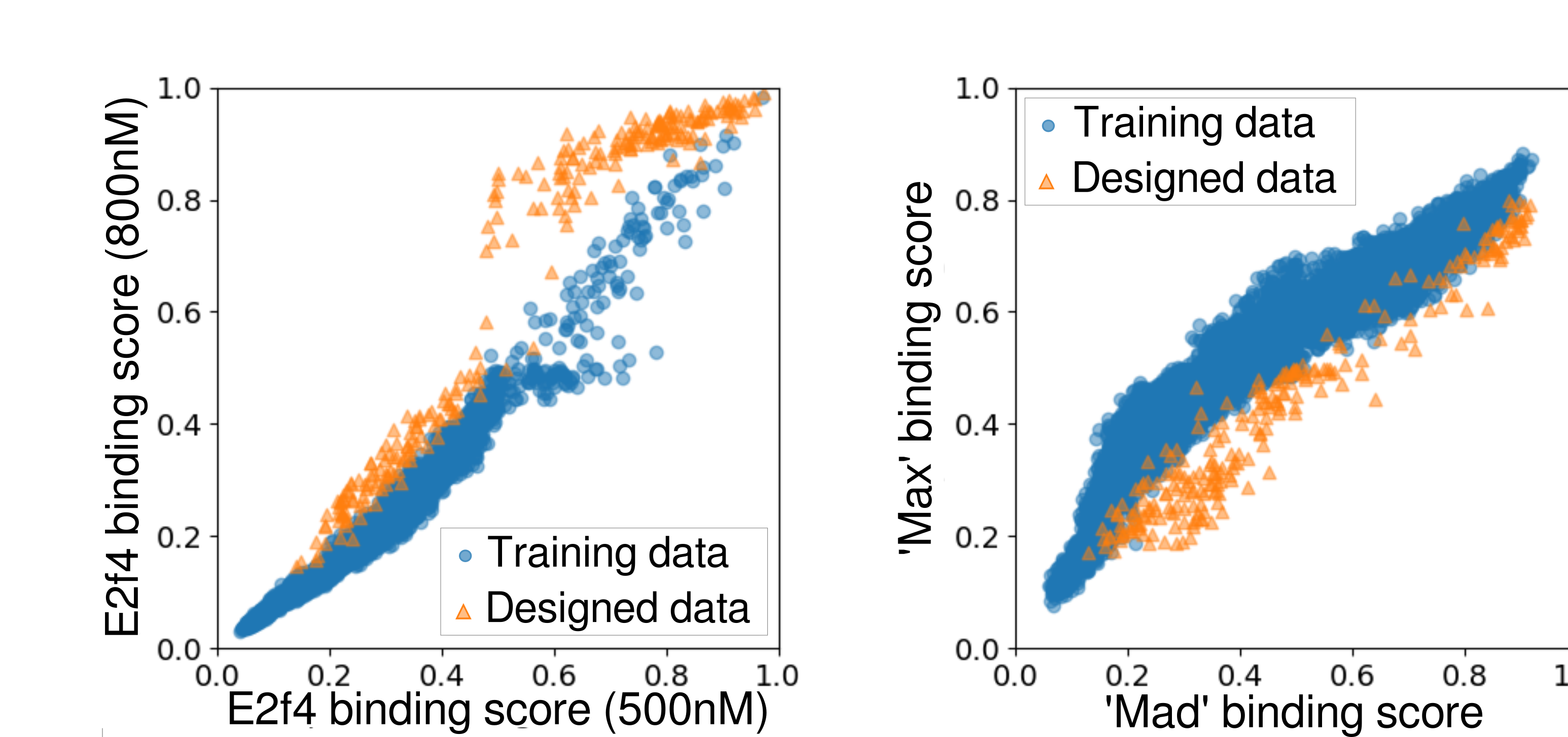}
    \caption{Designing DNA sequences to trade off multiple properties at once, demonstrated with the proteins Mad, Max, and E2f4. We are able to discover sequences which push the boundaries of what was found in the training data. Note: for best results, we weighted the Mad predictor with a factor of 1.2 relative to the Max predictor, while both E2f4 costs had the same weight (cf. Eq. (\ref{eq:multi_objective})).}\label{fig:multi_scatter}
\end{figure*}

\paragraph{Design process} Our protein-binding dataset contains binding measurements on the same probe sequences for multiple proteins from the same family. Leveraging this, our goal is the following: to design DNA sequences which preferentially bind to one protein in a family but not the other. We also undertake this challenge for the situation where the two predictors model binding of the same protein, but under two different molecular concentrations. Sample results of this design process are shown in Fig. \ref{fig:multi_scatter}. Like in Sec. \ref{sec:learned_predictor}, we are able to design many sequences with characteristics that generalize well beyond the explicit content of the training data. Because of the underlying similarities, the two predictors largely capture the same structure, differing only in subtle ways. Our design process lets us explore these subtle differences by generating sequences which exhibit them.

\section{Summary \& Future Work}

We have introduced several ways to generate and design genomic sequences using deep generative models. 
We presented a GAN-based generative model for DNA, proposed a variant of activation maximization for DNA sequence data, and combined these two methods together into a joint method. 
Our computational experiments indicate that these generative tools learn important structure from DNA sequences, even from restricted information, and can be used to explore and design new DNA sequences with desired properties.

\paragraph{Experimental validation}
The framework outlined here should be seen as a new set of tools for machine-aided sequence design, but they may need to be further improved and fine-tuned with suitable wet lab experiments before being put into real use. This is especially true for critical, real-world applications, such as in synthetic biology and genome editing, since even a single-nucleotide variation (SNV) can have dramatic consequences for a living organism. Ultimately, cellular machinery is the judge of the quality and utility of a designed sequence.

\paragraph{Drawbacks and limitations}
While we found these techniques to generally be well-behaved, we did notice some undesirable cases during our explorations. For example, in Sec. \ref{sec:motif_matching}, we sought to generate sequences where a particular motif appeared. We observed that sometimes a strong motif match would appear twice in the same generated sequence, e.g., the second sequence of Fig. \ref{fig:motif_matching}. This occurred because the final predictor score was insensitive to the presence of two motifs (the score is based on whichever one is the best match). Biologically, however, such sequences could be undesirable. 

One method to discourage known undesirable outcomes is to add extra terms in the overall predictor objective, Eq. (\ref{eq:multi_objective}), which penalize badly behaved sequences (for the above failure, we would want to penalize more than one strong motif match). Such penalties can be incorporated straightforwardly in any situation where we want to suppress known undesirable behaviour of the generated sequences. On the other hand, we must remember that while there will be some constraints which are `known unknowns,' there will also be `unknown unknowns.' The former are constraints which we know (or suspect) are there, but we are uncertain about how to quantify; the latter are constraints whose existence we are not even aware of given our current understanding of genomics. Using a generator model learned from real data can help to offset, if not completely alleviate, such worries.

\paragraph{Future machine learning directions}
We would like to emphasize that generative neural network tools are still at a relatively early development stage. Even in the image domain, where generative methods have shown the most striking performance, we are at the moment still below the threshold where machine-generated and real data are consistently indistinguishable. In natural language, the gap between generated and authentic data is even wider. Despite these facts, there is still huge enthusiasm and high expectations for generative models in the deep learning community, and a great degree of research is currently being undertaken in this area. 

With this paper, we hope to expand this enthusiasm to the genomics domain. This work should be seen as the first detailed exploration of generative DNA design in this subject area, and the experiments presented here are the initial validation and proofs-of-concept of this new research direction. Like the aforementioned domains, we do not expect the generated data to be completely flawless. However, the design architecture presented here is constructed to be easily upgradable as new progress is made in generative models and algorithms, and as genomics datasets get larger, richer, and more accurate. Any improvements in these areas will only enrich and enhance the capabilities of our presented frameworks.

There are also a number of interesting directions to explore. One next step would be to train an encoder $E$ which maps data back to latent codes: $E(\mathbf{x})=\mathbf{z}$, making it easier to find latent encodings for specific sequences. Another possible extension could be to build a conditional GAN model and perhaps combine it with the joint architecture; this would allow some properties to remain fixed while others were tuned. Finally, we could consider adapting recent methods \cite{isola2016image, zhu2017unpaired} where data is translated from one domain to another (e.g., aerial images to maps, daytime to nighttime images, horse photos to zebra photos) to genomics sequences. For instance, we could provide a `map' of where we want certain components (introns, exons, promoters, enhancers) to be, and a generative model would dream up plausible sequences with the desired properties.

\paragraph{Future biological directions}

On the biological side, the door is now open to use these ideas in practice. 
We have proposed and explored several potential biological applications. 
A direction of future work is to combine these approaches with an experimental validation stage to confirm the properties of designed sequences. 
We foresee an active learning setup, similar to Sec. \ref{sec:learned_predictor}, where a predictor is optimized to generate a small number of high-value sequences which push the limits of its predictive capabilities. 
These sequences can then be measured experimentally and the resulting data used to retrain and improve the original predictor.

\subsubsection*{Acknowledgments}

We thank NVIDIA for providing GPUs used for this work. All sequence figures were created with WebLogo \cite{crooks2004weblogo}.

\bibliographystyle{unsrt}
\bibliography{references}

\clearpage
\appendix

\section{Further Experiments on the Latent Space}
\label{app:complementation}

In the main text, we showed that a WGAN trained on a dataset of DNA sequences taken from chromosome 1 of the human genome learned the notion of DNA complementation and encoded this using a reflection in the latent space. In this section, we flesh out this discovery in more detail. 

\begin{figure}[t]
 \centering\includegraphics[width=\textwidth]{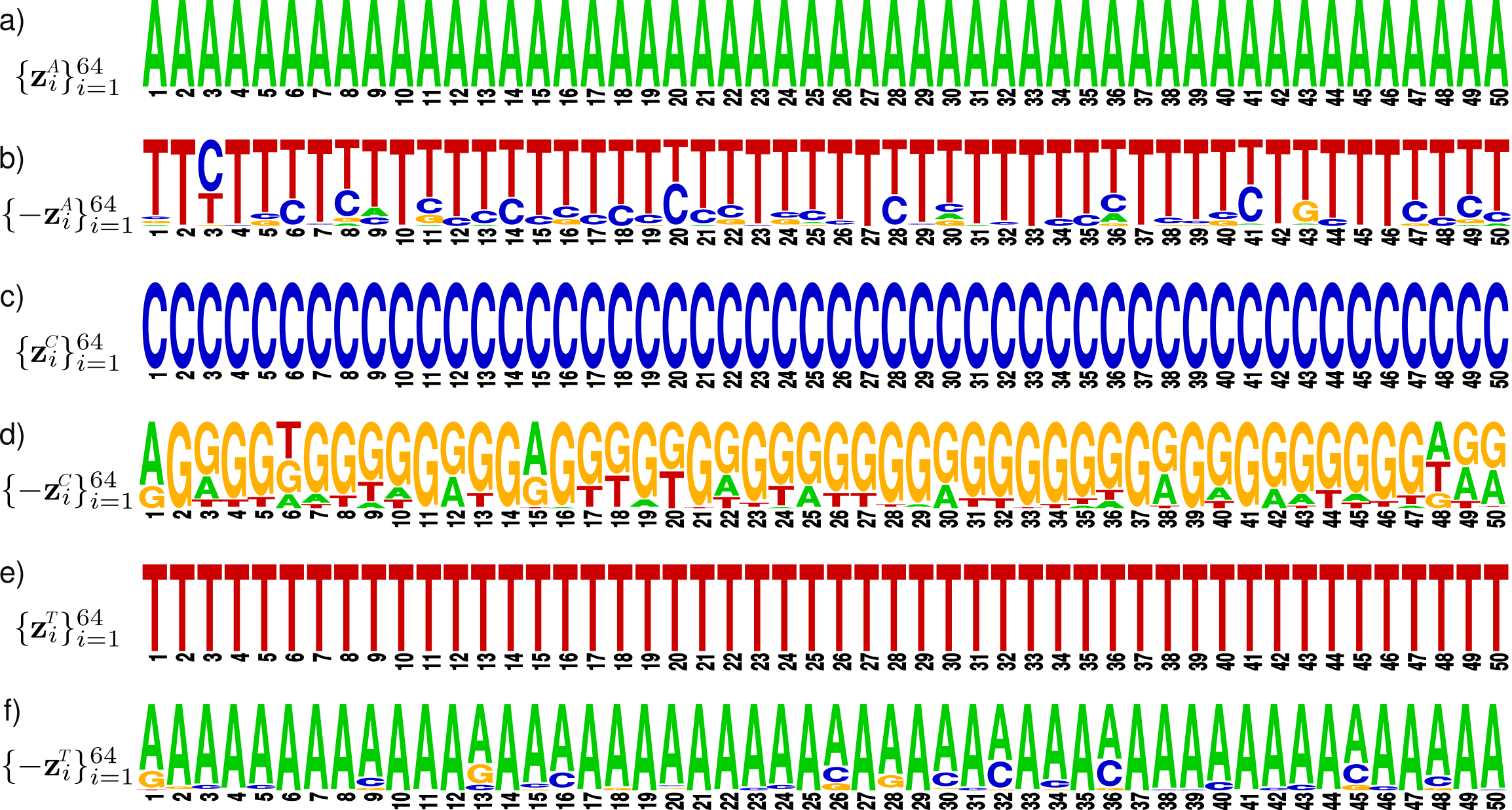}
 \caption{Latent space complementation with a batch of points $\{\mathbf{z}^N_i\}_{i=1}^{64}$ for $N=A,C,T$. 
 Parts a), c), e): Each latent code in the batch generates the same sequence, containing one repeated nucleotide. 
 Parts b), d), f): The reflections of the latent codes from parts a), c), e), respectively, generate sequences strongly biased towards the complementary nucleotides. Letter heights reflect their relative frequency in the batch at each position.}\label{fig:latent_complementation_ACT_short}
\end{figure}

In Fig. \ref{fig:latent_complementation} of the main text, we considered a batch of latent vectors $\{\mathbf{z}_i\}_{i=1}^{64}$ which each generated the sequence containing only the nucleotide G. Reflecting these latent points, the generated sequences then showed a strong bias towards the complementary nucleotide C. Here we repeat this experiment -- with the same generator model -- using each of the three other nucleotides for the starting sequence; the results are shown in Fig. \ref{fig:latent_complementation_ACT_short}. In each case, we can see a clear bias towards the complementary nucleotide A$\leftrightarrow$T, C$\leftrightarrow$G. 

To rule out that the correct complementation behaviour occurred merely by chance, we trained a new WGAN generator for each of the 24 possible encodings of the nucleotides \{A, C, G, T\} as separate one-hot vectors $\{[1,0,0,0]^T, [0,1,0,0]^T, [0,0,1,0]^T, [0,0,0,1]^T\}$.
We redid the reflection experiment for each of these models, with A being the repeated nucleotide. In 22 of these models, a strong pairing of A with the complementary base T via latent space reflection was observed. 

For the two cases where an obvious complementation behaviour was not observed, we trained those models a second time from scratch with the exact same architecture, differing only in the values of random parameter initializations and random noise values used during training. Both of the retrained models manifested the desired complementation/reflection correspondance. This suggests that the observed complementation behaviour in models is not dependent on the particular one-hot encodings we use. Rather, it is something that is inherent in the distribution of the data. We hypothesize that the two outlier models did not have a strong complementation/reflection correspondance because, compared to the other 22 models, they fell into an alternative parameter regime during training.

We also observed that the complementation/reflection correspondance was primarily evident for sequences with the same character repeated multiple times. Sequences without such regular structure presented no obvious degree of complementation between $\mathbf{z}$ and $\mathbf{-z}$. This indicates that the model encodes approximate complementatarity only for low-entropy sequences.

\section{Model and Dataset Details}
\subsection{Models}
\label{app:models}

\begin{figure}[t]
 \includegraphics[width=\textwidth]{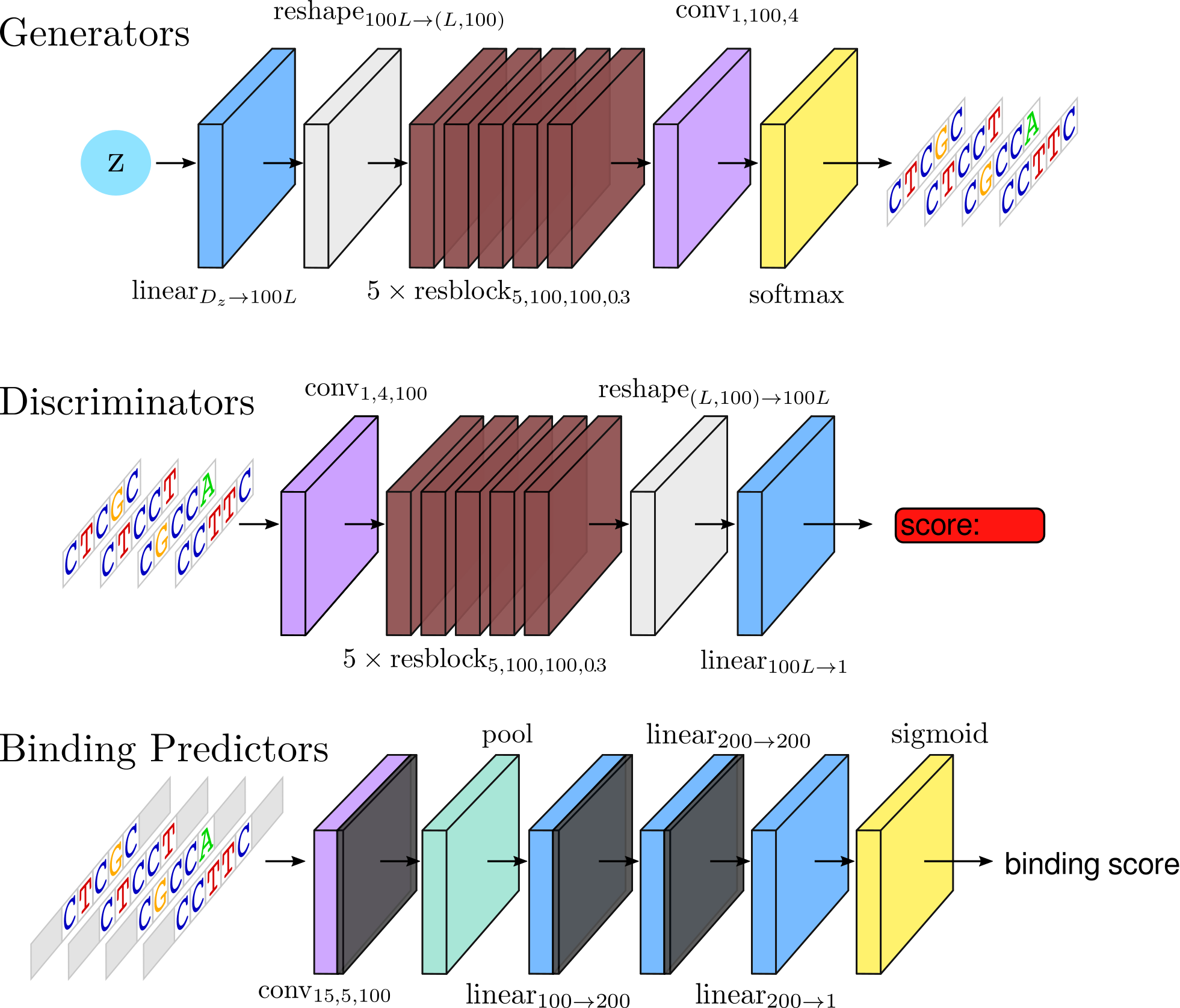}
 \caption{Architectures for the different models used in this paper. Gray flanking areas on input sequences represent padding and thin dark layers represent leaky relu activations. $D_\mathbf{z}$ is the dimension of the latent space, while $L$ denotes the length of DNA sequences.}\label{fig:model_architectures}
\end{figure}

For simplicity, we use the following abbreviations for model layers:
\begin{itemize}
 \item $\mathrm{linear}_{D_\mathrm{in}\rightarrow D_\mathrm{out}}$: linear layer, consisting of a multiplication by a weight matrix of shape $(D_\mathrm{in}, D_\mathrm{out})$ and addition of a bias parameter of dimension $D_\mathrm{out}$
 \item $\mathrm{conv}_{L, N_\mathrm{in}, N_\mathrm{out}}$: one-dimensional convolution operation with filters of length $L$, mapping from $N_\mathrm{in}$ channels to $N_\mathrm{out}$ channels, followed by addition of a bias parameter of dimension $N_\mathrm{out}$
 \item $\mathrm{resblock}_{L, N_\mathrm{in}, N_\mathrm{out}, r}$: a residual block with 2 internal layers, each layer consisting of a rectified linear unit (relu) followed by a $\mathrm{conv}_{L, N_\mathrm{in}, N_\mathrm{out}}$ transformation. The output of the second layer is multiplied by $r\leq 1$ and added back to the resblock input
 \item $\mathrm{reshape}_{S_1\rightarrow S_2}$: change shape of features from $S_1$ to $S_2$
 \item $\mathrm{pool}$: max pooling and average pooling of activations within each feature, then concatenation of these two pooling outputs
 \item $\mathrm{softmax}$: final activation function is a softmax over 4 dimensions (representing the 4 nucleotides), applied separately to each position along the sequence length
 \item $\mathrm{sigmoid}$: final activation function is a sigmoid function
\end{itemize}
In practice, all operations are performed with batches of data. For simplicity, we have not denoted the extra batch dimension in these operations. The learnable model parameters consist of the weights, the biases, and the convolutional filters. 

\paragraph{Generators}

Every experiment in this paper involving a generator model used the Wasserstein GAN architecture trained with a gradient penalty (`WGAN-GP'), based on the WGAN language model from \cite{gulrajani2017improved} (code for that paper is available at \url{https://github.com/igul222/improved_wgan_training}). Specifically, these models use a residual architecture \cite{he2016deep} in both the generator and discriminator. For sequences of length $L$, the WGAN models used are shown in Fig. \ref{fig:model_architectures}. The dimension of the latent space used for all models was $D_z=100$. During training, we sampled from the latent space according to a standard normal distribution, and we performed 5 discriminator updates for every generator update.

\paragraph{Predictors}

For several of our computational experiments, we made use of learned networks for predicting DNA-protein binding. These networks all have a convolutional neural network architecture (shown in Fig. \ref{fig:model_architectures}) similar to the DeepBind model introduced in \cite{alipanahi2015predicting}. The oracle model used in Sec. \ref{sec:learned_predictor} also has this architecture. To allow partial sequence matches, the incoming data is padded on both left and right, with each nucleotide channel having the value of $0.25$. 

In order for our design methods to work properly, we found that it was important to replace standard relu units with a `leaky' relu, given by $\mathrm{leaky\_relu}(\mathbf{x}) = \max(\alpha \mathbf{x}, \mathbf{x})$, where we used $\alpha=0.1$. This ensures that there is nonvanishing gradient information available for almost every value of $\mathbf{x}$. 


\paragraph{Joint Models}

We already outlined the main calculations involved to design synthetic data using the joint framework in Sec. \ref{sec:plug_and_play}. Here we want to briefly mention that this framework also has the option to use additional terms in the latent space update, Eq. (\ref{eq:deepdream_step_z}). The first term we could add corresponds to a prior on the latent space $\epsilon_2\nabla_{\mathbf{z}}\log p(\mathbf{z})$, where $\epsilon_2$ is a user-specified hyperparameter. This can be seen as an extra constraint forcing designed sequences to remain closer to those latent values which were sampled during the generator's training. In the experiments presented in this paper, we did not make use of the prior term. The second optional component for Eq. (\ref{eq:deepdream_step_z}) is to add random noise. This can help to avoid local optima and potentially encourage diversity during the design phase. In our experiments, we sampled from a normal distribution with zero variance and standard deviation $10^{-5}$. Optimization during the design phase was carried out using the Adam optimizer \cite{kingma2014adam} with a step size of $10^{-1}$ and the suggested default values for the decay parameters, $\beta_1=0.9$, $\beta_2=0.999$.

\subsection{Datasets}
\label{app:datasets}

\paragraph{chr1} This dataset was compiled from chromosome 1 (the largest chromosome) of the human genome \cite{lander2001initial} (Dec. 2013) hg38 assembly. The data was downloaded as the fasta-format file chr1.fa from the UCSC genome browser \cite{kent2002human}, available at the website \url{http://genome.ucsc.edu}. This file contained nucleotide sequences of length 50. We processed this data in two steps: i) all characters were converted to uppercase, and (ii) sequences containing the character N (a placeholder representing an unknown nucleotide) were filtered out. We used 80\% of this filtered data to make a training set, with the remaining 20\% split evenly into validation and test sets.

\paragraph{exons-50-400-500} This dataset was compiled from the human genome (Dec. 2013) hg38 assembly by cross-referencing with the `knownCanonical' table of canonical splice variants of genes (`+' strand only) in the UCSC genome browser. Specifically, we filtered the entire human genome for exon sequences with lengths between 50-400 nucleotides. Flanking sequences were also captured for each exon, such that each sequence contained a total of 500 nucleotides, with the exon appearing at the centre. This data was filtered to remove sequences with the placeholder nucleotide N, as well as those where the the captured flanking sequences were less than 50 nucleotides on either side. We used 80\% of this data for a training set, with validation and test sets containing 10\% each. Along with this exonic sequence dataset, a complementary annotation file was created to identify the location of exons within each sequence in the dataset. For each sequence, the annotation track has a value of 1 at positions which are part of the exon, and zero otherwise. Models were trained with the combined sequence/annotation data.

\paragraph{protein-binding} This dataset was provided to us courtesy of \cite{raluca2017}. It consists of 36-nucleotide-long probe sequences along with experimentally measured binding values for these sequences. The data is split up into 4 subsets, each corresponding to a particular protein family. In some families, we also have measurement data for a single protein at multiple concentrations. Within a family, all proteins/concentrations are scored on the same sequences. The probe sequences are specially designed to cover a broad range of binding values (normalized between 0 and 1). The number of available sequences varied between 7.1-25.4k, depending on the family. In all experiments we used a 80:10:10 train/valid/test split.

\section{Generative Modeling Architectures}
\label{app:gen_archs}

In this section, we will briefly outline the most competitive generative modeling architectures (see Fig. \ref{fig:gen_models}) and discuss some observations as to their suitability for modeling and tuning genomics sequences.

\subsection{Recurrent Neural Networks}

\begin{figure}[t]
 \includegraphics[width=\textwidth]{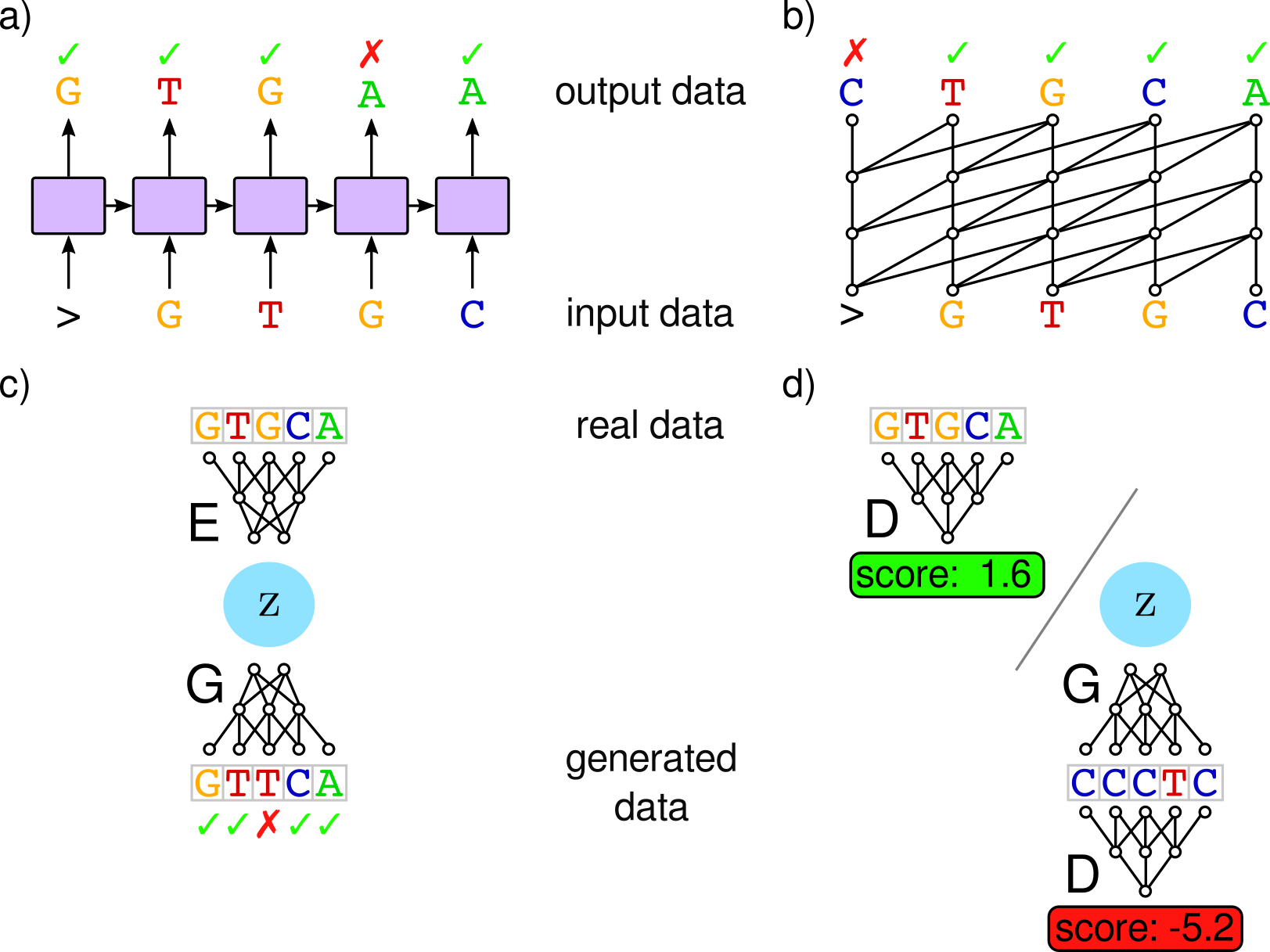}
 \caption{Generative neural network models shown with short example sequences: 
 a) Recurrent neural network; 
 b) PixelCNN; 
 c) Variational Autoencoder; 
 d) Generative Adversarial Network. A generic starting character (e.g., `>`) is used to prompt the RNN and PixelCNN at the first time step.}\label{fig:gen_models}
\end{figure}

RNNs (Fig. \ref{fig:gen_models}a) are a natural choice when working with sequences, since they process data in a sequential manner. RNNs consist of a single computational cell which sequentially processes inputs and produces outputs while retaining some internal state $\mathbf{h}$. Concretely, at time step $t$, the network takes input data $\mathbf{x}_t$, and processes it in combination with its previous internal state $\mathbf{h}_{t-1}$ using a feedforward network, producing an output $\mathbf{o}_t =\mathbf{o}(\mathbf{x}_t, \mathbf{h}_{t-1})$ and an updated internal state $\mathbf{h}_{t}=\mathbf{h}(\mathbf{x}_0, \mathbf{h}_{t-1})$. Commonly employed RNN cells are the Long Short Term Memory (LSTM) \cite{hochreiter1997long, graves2013generating} and the Gated Recurrent Unit (GRU) \cite{cho2014properties, chung2014empirical}.

To train an RNN model, one usually feeds data from the training set as input -- say $\mathbf{x}_t$ at time $t$ -- and treats the output $\mathbf{o}_t$ as a probability distribution representing the model's prediction for the next character in the sequence $\mathbf{x}_{t+1}$ (at the first time step, we feed the model some data-independent `start` character). We train the model to maximize the likelihood of predicting the correct next character. To use an RNN as a generative model, we simply sample its prediction at each time $t$, then feed this back as the input at the next step $t+1$, generating arbitrarily long sequences. RNNs can also be trained to generate sequences in conditional manner, producing outputs which have some desired property. We can do this by appending extra labelled data $\mathbf{y}$ (which may signal, for example, the language or overall sentiment of the full sequence) to the inputs $\mathbf{x}_t$.

\paragraph{Suitability for DNA Design} To our knowledge, there is no successful variant of the activation maximization or plug \& play methods that operates on RNNs. This leaves us with the vanilla RNN generative mode described above as currently the only viable option for DNA design. Such models could certainly be sufficient for some genomics applications. If we want to tune or optimize DNA sequences, there is even the option to train conditional RNN models, provided we have enough supervised data. Of course, without a learned latent encoding, we are limited to tune those properties for which we explicitly trained the conditional model for. 
We note that Refs. \cite{karpathy2015unreasonable, radford2017learning} demonstrated that one can sometimes inspect the activations of RNNs trained in an unsupervised manner and find neuron units whose activation captures some property of interest in the sequence. However, this inspection process could be quite time-consuming, with no guarantee that we will uncover the specific properties that we are actually interested in.

\subsection{Deep autoregressive models}
As generative models, RNNs are autoregressive, meaning they generate outputs at time $t={N+1}$ by conditioning on their previous outputs from $t={0,...,N}$. For RNNs, this past history must be captured via the internal state $\mathbf{h}_t$, and popular RNN cells are explicitly designed to permit long memories. Refs. \cite{oord2016pixel, van2016conditional} demonstrated that instead of feeding inputs only one at a time and relying on the network to memorize past inputs, we can alternatively show it the entire past history up to that point. This frees up the network from having to memorize and it can devote all its resources to processing. 

With deep autoregressive models, this processing of the past inputs is achieved with a deep neural network, which could be an RNN or a convolutional neural network (see Fig. \ref{fig:gen_models}b). Analogously to recurrent networks, such models can be used to generate sequences by feeding their own history of previous predictions as input for each time step. These models can also be built as conditional models, enabling the generation of sequences with tailored properties. The first application area for these models was in images, so they were called PixelRNN and PixelCNN. More recently, similar ideas have been successfully applied to language \cite{kalchbrenner2016neural} and speech \cite{van2016wavenet}.

\paragraph{Suitability for DNA Design} In general, we expect these models to be more powerful than recurrent networks, simply because they get to see much more past history for context. Yet, for design, they have the same problems as recurrent networks, namely that all properties of interest will require supervised training with a labelled dataset and that these properties must be chosen beforehand and built in during training. 

\subsection{Variational Autoencoders}

In contrast to the two models above, VAEs (Fig. \ref{fig:gen_models}c) have the ability to learn a controllable latent representation of data in an \emph{unsupervised} manner. By changing the latent variable $\mathbf{z}$, we can modify the synthetic data that the model generates. In principle, this could provide us with a lot of power and flexibilty for tuning generated data.

Autoencoder models learn two transformations: an encoder $E$ transforming data to latent variables, $\mathbf{x}\rightarrow\mathbf{z}$, and a decoder (or generator) $G$ transforming latent variables to generated data, $\mathbf{z}\rightarrow\mathbf{x}'$. VAEs use probability distributions rather than deterministic functions to model these transformations. To encode, we sample $\mathbf{z}$ from a distribution $q(\mathbf{z}|\mathbf{x})$; to decode, we do likewise for $\mathbf{x}$ from a distribution $p(\mathbf{x}|\mathbf{z})$. Both these distributions are modelled via neural networks. To use a trained VAE as a generative model, we only need to use the decoder network.

When training autoencoders, the goal is to make the error from the encoding/decoding process $\mathbf{x}\rightarrow\mathbf{z}\rightarrow\mathbf{x}'$ as small as possible. For VAEs, this reconstruction error is given by
\begin{equation}
 \mathcal{L}_\mathrm{recon}:=\mathbb{E}_{\mathbf{z}\sim q(\mathbf{z}|\mathbf{x})}[-\log p(\mathbf{x}|\mathbf{z})].
\end{equation}
In order to reconstruct successfully, the model must learn how to capture the essential properties of the data within the latent variable $\mathbf{z}$. VAEs also include regularization which encourages the latent codes to vary smoothly. This is captured by a KL divergence term between $q(\mathbf{z}|\mathbf{x})$ and a fixed prior distribution on the latent space $p(\mathbf{z})$ -- commonly a standard normal. The full VAE objective which is minimized during training is
\begin{equation}
 \mathcal{L}_\mathrm{recon} + D_\mathrm{KL}(q(\mathbf{z}|\mathbf{x})|| p(\mathbf{z})).
\end{equation}
 
\paragraph{Suitability for DNA Design} While VAEs offer tantalizing potential as generative models, they have some inherent issues that make them tricky to use on sequence data.  It has been observed \cite{bowman2015generating,chen2016variational} that if we use a strong decoder network, such as an RNN, VAEs will exhibit a preference to push the KL divergence term to zero. This causes the latent code to be ignored and the generative process is handled completely by the decoder. Without learning a meaningful latent code, such models are no better than a standard RNN. Some methods have been explored to remedy this problem, such as turning off the KL regularization term early in training \cite{bowman2015generating} or restricting the modeling power of the decoder to improve the quality of the latent representation \cite{chen2016variational, yang2017improved}.

Generally, we found VAEs unreliable and poorly suited for generative modeling of real-world genomics data. We tested a number of VAE models where both the encoder and decoder were RNNs. While these models performed well for simple toy scenarios, once we graduated to more realistic data, their performance suffered. In particular, on datasets with a large amount of noise (e.g., a single DNA motif within a larger noisy background sequence), the VAE+
RNN models converged to have near-zero KL divergence, effectively ignoring the latent codes and modeling the noisy sequence with just the RNN decoder. This behaviour persisted even when we removed the KL term for the first stages of training. We hypothesize that the high degree of noise made it difficult for the models to properly autoencode data. 

Another potential issue is that VAEs model the generative process stochastically (via a distribution $p(\mathbf{x}|\mathbf{z})$). Because of this, a fixed latent code might decode to multiple generated samples. This leads to the question of how much of the variation in the generated data is accounted for by the latent code $\mathbf{z}$ and how much by the stochastic process $\mathbf{x}\sim p(\mathbf{x}|\mathbf{z})$. While not a huge stumbling block, this may interfere with our goal of controlling and fine-tuning the outputs. In effect, we give up some control of the generated data by having a nondeterministic generator.

\subsection{Generative Adversarial Networks}

\paragraph{Suitability for DNA Design} One advantage of GANs (Fig. \ref{fig:gen_models}d) is the deterministic generator; all variations in the generated data are accounted for by the latent code $\mathbf{z}$. As well, unlike VAEs, GANs are not known to ignore the latent code. Thus, the latent code forms a reliable high-level representation of data and we can modify and tune the generated sequences continuously simply by moving to different coordinates in the latent space. 

As a recent invention, GANs still have a number of kinks to smooth out. One major known issue with GANs is their instability during training. Because the generator and discriminator networks are being optimized in a back-and-forth manner, we can run into issues where one network dominates the other or where the two networks fall into an oscillatory training pattern, leading to clearly suboptimal generative models. Another known issue is that the generator will sometimes concentrate its samples on a single or small number of modes, effectively ignoring much of the variation in real data. The improved WGAN architecture presented in \cite{gulrajani2017improved} -- and used throughout this work -- has proven effective in overcoming many of these GAN instabilities.

\end{document}